# OpenAI Cribbed Our Tax Example, But Can GPT-4 Really Do Tax?

**by Andrew Blair-Stanek, Nils Holzenberger, and Benjamin Van Durme**

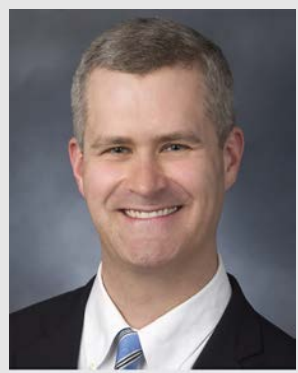
Andrew Blair-Stanek

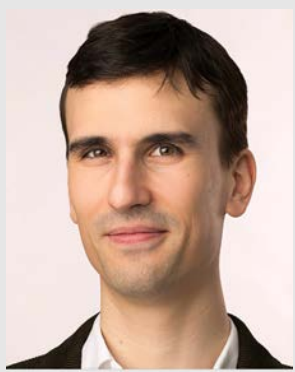
Nils Holzenberger

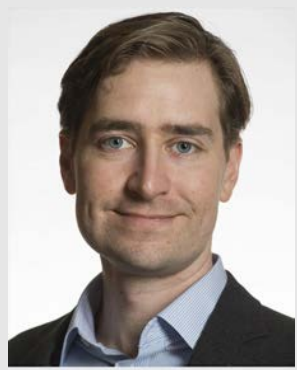
Benjamin Van Durme

Andrew Blair-Stanek is a professor of law at the University of Maryland; Nils Holzenberger is an associate professor in computer science at Telecom Paris, Institut Polytechnique de Paris; and Benjamin Van Durme is an associate professor in computer science at Johns Hopkins University. This work has been supported by the U.S. National Science Foundation under grant No. 2204926.

In this article, the authors explain where OpenAI got the tax law example in its livestream demonstration of GPT-4, why GPT-4 got the wrong answer, and how it fails to reliably calculate taxes.

When OpenAI debuted its GPT-4 AI language model in a March 14 livestream, it used a tax law example to demonstrate the model's power.[1] The presenter pasted in what he called "about 16 pages' worth of tax code"[2] and then seven sentences of facts about married couple Alice and Bob, who have a son Charlie and $36,991 and $41,990 of income, respectively.

These seven sentences about Alice, Bob, and Charlie come word-for-word from a handcrafted data set we developed at Johns Hopkins University and published in 2020 for training and measuring AI models for reasoning over statutory language.[3] Our data set has been freely available for download since 2020,[4] allowing OpenAI and others to use it independently of our research efforts. Every word, punctuation mark, and number in the taxpayer facts comes exactly from our tax_case_9 — even the percent sign at the start of the line.[5] (See Figure 1.)

---

[1] The entire livestream is available at OpenAI, "GPT-4 Developer Livestream," YouTube (Mar. 14, 2023), and has over 3 million views. The tax law example starts at minute 19:11.

[2] *Id.* at minute 19:51.

[3] Nils Holzenberger, Andrew Blair-Stanek, and Benjamin Van Durme, "A Dataset for Statutory Reasoning in Tax Law Entailment and Question Answering," Proceedings of the 2020 Natural Legal Language Processing Workshop (2020).

[4] Johns Hopkins Natural Language Processing for Law, StAtutory Reasoning Assessment (SARA), available at https://nlp.jhu.edu/law/sara/sara.tar.gz. Go to the directory "Cases" to find the file tax_case_9.pl. Line 2 was copied in the livestream.

[5] Tax_case_9.pl is written in the programming language Prolog. A percent mark indicates to Prolog that the line is a human-readable comment, not to be executed.

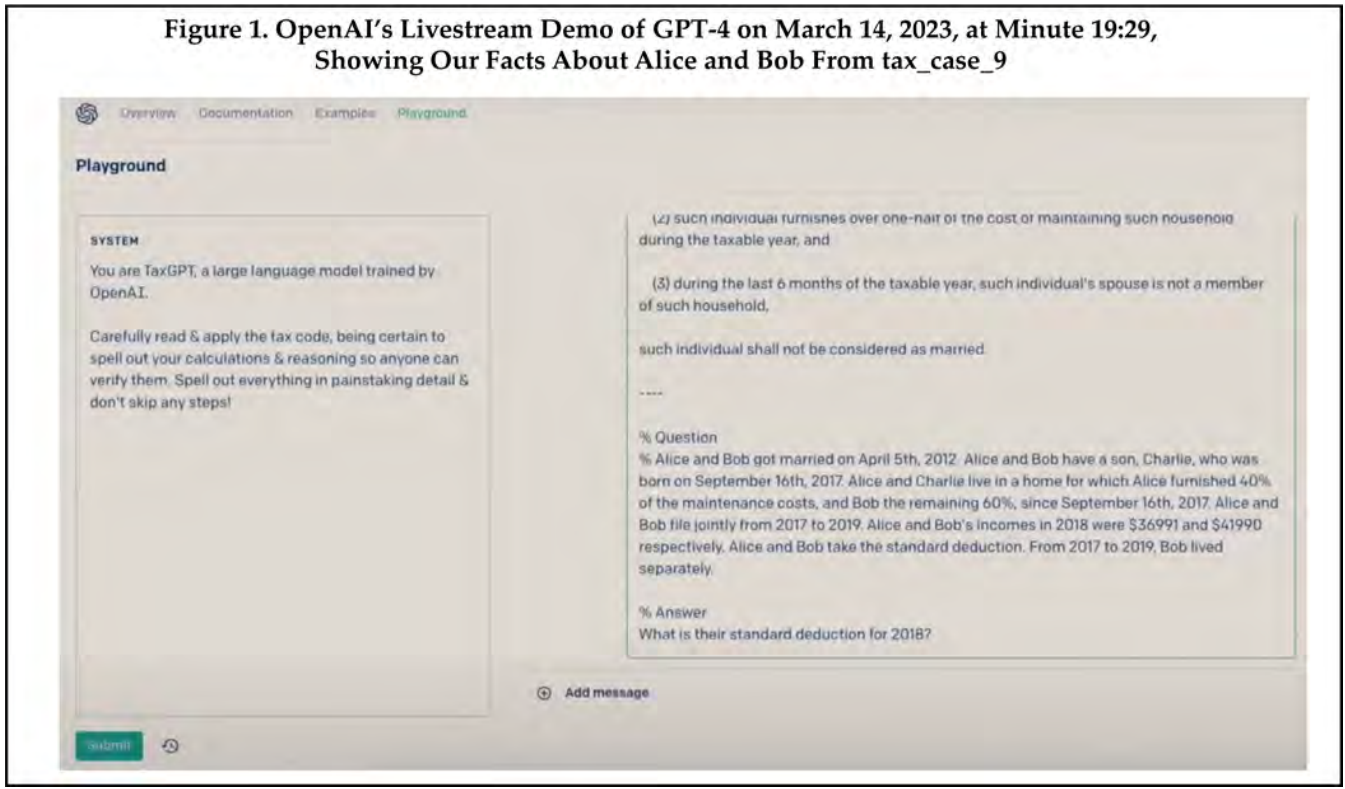

Figure 1. OpenAI's Livestream Demo of GPT-4 on March 14, 2023, at Minute 19:29, Showing Our Facts About Alice and Bob From tax_case_9

Where did the "about 16 pages' worth of tax code" come from? Again, from our 2020 data set. We called our 2020 data set SARA, an acronym for StAtutory Reasoning Assessment. SARA has two parts: statutes and cases. There are 376 handcrafted cases in SARA; tax_case_9 is one of them. The statutes consist of nine sections of the IRC,[6] heavily edited by hand to be much shorter and remove ambiguity.[7] If you put all the SARA statutes into a single file it will be about 16 pages long (depending on the font).[8]

One of our edits was paring section 1 down to only sections 1(a) through (d), which contain the Clinton-era tax rates. We cut section 1(j), which contains the reduced Tax Cuts and Jobs Act rates for 2018-2025. This editing explains why GPT-4 got the wrong answer on the livestream for Alice and Bob's 2018 taxes. We did not, however, edit out the TCJA standard deduction increase at section 63(c)(7), which explains why GPT-4 correctly found that Alice and Bob's standard deduction for 2018 was $24,000.[9]

From minute 20:07 to 20:40 of the livestream, we see some of the tax sections pasted into GPT-4. These are SARA's heavily edited version of the IRC — not actual IRC sections. For example, at 20:23, we see part of section 63(c) with the paragraphs jumping from (3) to (5); in SARA, we edited out (4). At 20:26, we see part of section 63(c)(6) with only subparagraphs (A), (B), and (D); in SARA, we edited out (C). At 20:40, we see parts of section 3306(b) with the paragraphs jumping from (2) to (7); in SARA, we edited out paragraphs (3) through (6). At 20:39 we see sections 3301 and 3306 regarding the federal unemployment tax; while these two sections are irrelevant to Alice and Bob's tax liability in tax_case_9, they are two of the nine sections we included in SARA. You can download our data set and compare it with the livestream's recording on YouTube.[10]

---

[6]Sections 1, 2, 63, 68, 151, 152, 3301, 3306, and 7703.

[7]The author Holzenberger did all the handcrafting and hand editing.

[8]"All_SARA_statutes.pdf," available at @BlairStanek/gpt-statues, Github.

[9]Livestream, *supra* note 1, at 20:42.

The presenter then gives directions to GPT-4: "Now calculate their total liability."[11] GPT-4 gives detailed step-by-step calculations and concludes that "Alice and Bob's total tax liability for 2018 is $10,597.68."[12] This is lauded as proof of GPT-4's amazing capabilities.

But this number is wrong, as Libin Zhang pointed out in *Tax Notes*.[13] Why did GPT-4 get it wrong? And why did OpenAI incorrectly think GPT-4 had gotten it right? In short: SARA.

Progress in artificial intelligence typically comes from creating and then solving simplified versions of more complex problems. That is why, for SARA, we limited ourselves to just nine IRC sections and substantially edited them down. We never intended SARA to be a real-world example for tax preparation, and our 2020 paper introducing SARA expressly warned:

> Because the statutes were simplified, the answers to the cases are not those that would be obtained with the current version of the IRC. Some of the IRC counterparts of the statutes in our dataset have been repealed, amended, or adjusted to reflect inflation.[14]

OpenAI got the facts about Alice, Bob, and Charlie from line 2 of the SARA file containing tax_case_9. On line 5 of the same file is the correct answer for their tax liability if the edited-down SARA statutes were the governing law — $10,598. That appears to be why OpenAI thought GPT-4 had correctly calculated the tax liability at $10,597.68. (We rounded to the nearest dollar for SARA, whereas GPT-4 calculates to the nearest cent.)

We empirically verified that using the SARA version of the IRC causes GPT-4 to get the wrong answer, exactly as in the livestream. First, we pasted into GPT-4 all nine SARA statutes, plus our facts about Alice, Bob, and Charlie.[15] Then we used the same "Now calculate their total liability" command. GPT-4 consistently answers $10,597.68, just as during the livestream.[16] Then we added section 1(j), which contains the TCJA rates applicable to 2018-2025, back into the SARA statutes. Aside from this one change, we reran the test just as it had been run in the livestream. GPT-4 correctly identified that section 1(j)'s rates supplanted section 1(a)'s rates in 2018 and came to the correct result — that Alice and Bob had a 2018 tax liability of $6,216.72. GPT-4 is impressive.

Our ultimate research goal is to develop "Shelter Check," an AI technology that will allow academic researchers, the IRS, Congress, and courts to proactively find tax minimization strategies before tax accountants and lawyers do. We first introduced the Shelter Check concept in the pages of *Tax Notes*,[17] and it has gotten some attention in the popular media.[18] Having AI models that can reason over tax statutes is a prerequisite for Shelter Check. We created the SARA data set to help train and measure AI models for statutory reasoning, so we are delighted that OpenAI used SARA and are excited about the potential for OpenAI's models to power Shelter Check one day.

SARA has 376 cases, of which tax_case_9 is just one, so we analyzed how GPT-4 performs on all of them. Each of the 376 cases has a series of facts, typically involving the taxpayers Alice and Bob, followed by a question. Of the 376 cases, 100 (of which tax_case_9 is one) have an answer that

---

[10]*See supra* note 4 for the download link. Go to the directory "Statutes," then the directory "Source" to find the edited IRC sections, each of which is a separate text file.

[11]Livestream, *supra* note 1, at 21:25.

[12]*Id.* at 21:50.

[13]Zhang, "Tax Questions for Language Models: A Follow-Up," *Tax Notes Federal*, June 5, 2023, p. 1699; *see also* Lauren Loricchio, "AI Programs Not Reliable Enough Yet for Tax Work, Attorney Says," *Tax Notes Federal*, June 26, 2023, p. 2241 (quoting Zhang).

[14]Holzenberger, Blair-Stanek, and Van Durme, *supra* note 3, at 2, column 2.

[15]We used model gpt-4-0314, with temperature = 1.0 and maximum length = 1200, as well as the same system prompt ("You are TaxGPT . . .") as the livestream, all to best reproduce the livestream.

[16]We ran it three times because of GPT-4 not being deterministic, especially with temperature = 1.0.

[17]Blair-Stanek, Holzenberger, and Van Durme, "Shelter Check: Proactively Finding Tax Minimization Strategies via AI," *Tax Notes Federal*, Dec. 12, 2022, p. 1515.

[18]"AI Has an 'Almost Game Changing Potential' to Close Tax Loopholes: Professor," Yahoo Finance, Apr. 14, 2023; Breck Dumas, "The AI Race to Find Tax Loopholes Is On," Fox Business, Apr. 19, 2023.

is a specific dollar figure. The remaining 276 have a binary answer of "true" or "false."

We posed all 276 true/false cases to GPT-4, pasting in the full SARA version of the IRC, plus the facts and questions.[19] GPT-4 got 186 correct, amounting to 67 percent accuracy. By comparison, a simple coin flip on true/false questions would result in 50 percent accuracy on average.[20] Let's consider a representative example of GPT-4 giving a wrong answer. Below is the prompt[21] we passed to GPT-4, with ellipses used so that the approximately 16 pages' worth of the SARA IRC is not all included:

> We are going to be doing reasoning applying the statutes below:
>
> . . .
>
> §151. Allowance of deductions for personal exemptions
>
> . . .
>
> (b) Taxpayer and spouse
>
> An exemption of the exemption amount for the taxpayer; and an additional exemption of the exemption amount for the spouse of the taxpayer if a joint return is not made by the taxpayer and his spouse, and if the spouse, for the calendar year in which the taxable year of the taxpayer begins, has no gross income and is not the dependent of another taxpayer.
>
> . . .
>
> Facts: Alice and Bob have been married since 2 Feb 2015. Bob has no income for 2015. Alice and Bob file their taxes jointly for 2015.

> Is the following statement True or False? Alice can receive an exemption for Bob under section 151(b) for the year 2015.

Here was GPT-4's response:

> True. According to section 151(b), Alice can receive an exemption for Bob under section 151(b) for the year 2015 since they are married, file their taxes jointly, and Bob has no gross income for 2015.

This is incorrect. Alice and Bob file jointly, and section 151(b) says a taxpayer gets an exemption for their spouse "if a joint return is *not* made by the taxpayer and his spouse." (Emphasis added.) Alice cannot receive an exemption for Bob under section 151(b). GPT-4 can make mistakes in reading statutes.

These mistakes do not surprise us. In a recent computer science publication,[22] we wrote a program to write simple "synthetic" statutes providing definitions of nonsense words. We then called GPT-3 text-davinci-003 — OpenAI's prior best model — to answer simple questions about these synthetic statutes. Because these synthetic statutes were entirely novel — they had never appeared before anywhere — they provided an unbiased measure of GPT-3's ability to reason over statutes. GPT-3 performed poorly compared with what we would expect from a human.[23]

The remaining 100 SARA cases involve calculating tax liabilities. We posed all 100 to GPT-4 in the same manner as the livestream, with the SARA IRC sections, followed by the facts and the question. Of the 100, GPT-4 refused to answer two, explaining that it needed more information. Of the remaining 98, it performed modestly well, getting 76 cases within 10 percent of the correct tax liability, 46 within 1 percent of the correct tax liability, and 31 within $1 of the correct tax liability. One of these 31 cases is tax_case_9. It seems that GPT-4's performance on tax_case_9 is not representative, because less than one-third of SARA cases were within $1. Figure 2 is a histogram showing the distribution of GPT-4's errors.

---

[19] We used model gpt-4-0314, with temperature = 0 to maximize reproducibility, as well as the same system prompt ("You are TaxGPT . . ."). We used the slightly updated version 2 of SARA: Johns Hopkins Natural Language Processing for Law, SARA v2, available for download at https://nlp.jhu.edu/law/sara_v2/sara_v2.tar.gz.

[20] These results are substantially better than the best results pre-GPT-3, published in Holzenberger and Van Durme, "Factoring Statutory Reasoning as Language Understanding Challenges," in Proceedings of the 59th Annual Meeting of the Association for Computational Linguistics (2021).

[21] This prompt is the "user" prompt; for the "system" prompt we used the same one as the livestream ("You are TaxGPT . . .").

[22] Blair-Stanek, Holzenberger, and Van Durme, "Can GPT-3 Perform Statutory Reasoning?" ArXiv (2023).

[23] *Id*. at Figure 1 and Table 4.

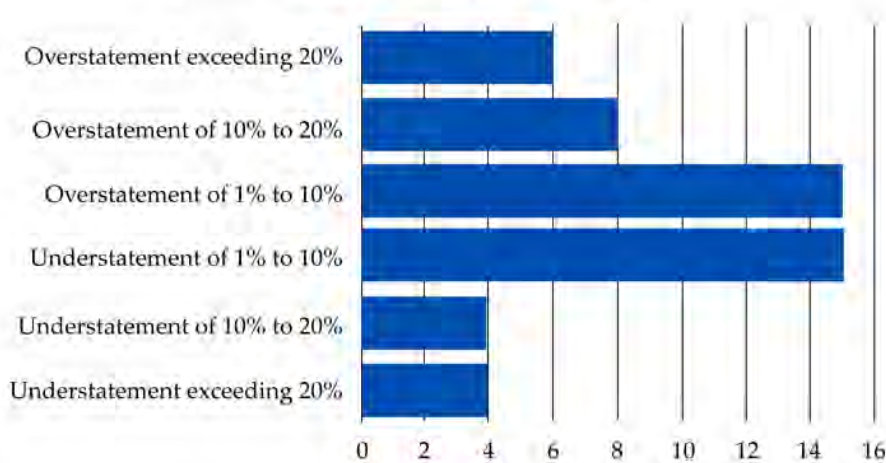

Figure 2. Histogram of Errors on the 52 SARA Tax Liability Questions Where GPT-4 Was Off by Over 1%

For the 22 cases in which GPT-4's answer was off by at least 10 percent, we reviewed GPT-4's reasoning to find the error. Of these 22 cases, 16 had just one error, four had two errors, and two had three errors. Notably, not one of the errors was mathematical; all involved misreading the statutes. The most common error, appearing in 10 cases, was the misreading of section 151 to provide for no personal exemption for years between 2010 and 2017, whereas personal exemptions are zeroed out only for 2018-2025. In seven cases GPT-4 misread section 63 to get the wrong standard deduction based on the taxpayer's filing status and the relevant year. Interestingly, in two cases GPT-4 simply ignored the federal unemployment tax imposed by section 3301, even though the facts clearly involved Alice paying wages subject to FUTA.

None of GPT-4's errors in these 22 cases resulted from ambiguities in the statutes. While many statutes have ambiguity, we specifically designed the SARA cases and edited the IRC sections to avoid any possible ambiguity. So statutory ambiguity explains none of GPT-4's errors.

GPT-4 is a remarkable model, able to take raw tax statutes and facts and correctly calculate the tax liability around one-third of the time — a large advance over what we thought possible just a few years ago. There is a large community of computer scientists working to expand the usefulness and power of these models. We are optimistic that future models will be able to help proactively find tax minimization strategies.

### Conclusion

In the livestream introducing GPT-4, OpenAI used one of our SARA tax cases verbatim, describing it as a real tax example, even though SARA is a simplified academic data set. In the demo, OpenAI also used our heavily edited SARA version of the IRC. OpenAI incorrectly thought GPT-4 had correctly calculated the tax liability because its answer matched the SARA answer, although our IRC edits change the result from the actual IRC. We tested GPT-4 on the entire SARA data set. It gets tax liabilities exactly right around one-third of the time and miscalculates tax liabilities by over 10 percent nearly a quarter of the time. GPT-4 often misreads even our simplified version of the IRC. In the livestream, the presenter warned, "You should always check with your tax adviser."[24] Wise advice. ∎

---

[24] Livestream, *supra* note 1, at 19:20.